\documentclass[journal]{IEEEtran}
\usepackage{amsmath,amsfonts}
\usepackage{algorithmic}
\usepackage{algorithm}
\usepackage{array}
\usepackage[caption=false,font=normalsize,labelfont=sf,textfont=sf]{subfig}
\usepackage{textcomp}
\usepackage{stfloats}
\usepackage{url}
\usepackage{verbatim}
\usepackage{graphicx}
\usepackage{cite}
\usepackage{lettrine}
\newcommand{\myparagraph}[1]{\vspace{0.1em}\noindent\textbf{#1}}
\usepackage[colorlinks,hyperindex,breaklinks]{hyperref}
\newcommand{\ie}{\textit{i}.\textit{e}.}
\newcommand{\eg}{\textit{e}.\textit{g}.}

\usepackage{pifont}
\newcommand{\cmark}{\ding{51}}%
\newcommand{\xmark}{\ding{55}}%
\usepackage{colortbl}
\usepackage{arydshln}
\usepackage{multirow}
\usepackage{multicol}
\usepackage{mathrsfs}
\usepackage{bm}
\usepackage{amsfonts}
\begin{document}
\title{Learning to Reduce Information Bottleneck for
Object Detection in Aerial Images}
\author{Yuchen~Shen, Dong Zhang, Zhihao Song, Xuesong Jiang, Qiaolin~Ye,~\IEEEmembership{Member,~IEEE}
\thanks{This work was supported by the Winter Olympic Science and Technology Service Project under Grant DA2020001, the National Science Foundation of China under Grant 62072246, and the Six Talent Peaks Projects of Jiangsu Province. (Corresponding author: Xuesong Jiang and Qiaolin Ye; Yuchen Shen and Zhihao Song contributed equally to this work.)

Y. Shen and Z. Song are with the College of Information Science and Technology, Nanjing Forestry University, Nanjing 210037, Jiangsu, China. E-mail: \{shenyuchen, songzhihao\}@njfu.edu.cn. 

Q. Ye is with the College of Information Science and Technology, Nanjing Forestry University, Nanjing 210037, Jiangsu, China, and also with Key Laboratory Intelligent Information Processing, Nanjing Xiaozhuang University, Nanjing 211171, Jiangsu, China. E-mail: yqlcom@njfu.edu.cn.

D. Zhang is with the Department of Computer Science and Engineering, The Hong Kong University of Science and Technology, Hong Kong, China. E-mail: dongz@ust.hk.

X. Jiang is with the College of Mechanical and Electronic Engineering, Nanjing Forestry University, Nanjing 210037, Jiangsu, China. E-mail: xsjiang@njfu.edu.cn.
}}
\markboth{submission of IEEE Geoscience and Remote Sensing Letters}
{Shell \MakeLowercase{\textit{et al.}}: A Sample Article Using IEEEtran.cls for IEEE Journals}
\maketitle
\begin{abstract} Object detection in aerial images is a fundamental research topic in the geoscience and remote sensing domain. However, the advanced approaches on this topic mainly focus on designing the elaborate backbones or head networks but ignore neck networks. In this letter, we first underline the importance of the neck network in object detection from the perspective of information bottleneck. Then, to alleviate the information deficiency problem in the current approaches, we propose a global semantic network (GSNet), which acts as a bridge from the backbone network to the head network in a bidirectional global pattern. Compared to the existing approaches, our model can capture the rich and enhanced image features with less computational costs. Besides, we further propose a feature fusion refinement module (FRM) for different levels of features, which are suffering from the problem of semantic gap in feature fusion. To demonstrate the effectiveness and efficiency of our approach, experiments are carried out on two challenging and representative aerial image datasets (\ie, DOTA and HRSC2016). Experimental results in terms of accuracy and complexity validate the superiority of our method. The code has been open-sourced at~\href{https://github.com/ssyc123/GSNet}{GSNet}.
\end{abstract}
\section{Introduction}
Object detection in aerial images is one of the most fundamental yet challenging research topics in the community of computer vision.
This topic aims at recognizing each object with a precise bounding box, 
which is the foundation of some potential application scenarios.
Recently, deep learning based object detection methods have made dramatic progresses~\cite{ding2019learning,liang2021learning,li2018deep}, thanks to the significant development of deep Convolutional Neural Networks (CNNs) on vision tasks, \eg, semantic segmentation~\cite{zhang2021self,zhang2020causal}, scene classification~\cite{zhang2019positional}. 

\begin{figure}[htp]
\centering
\includegraphics[width=0.48\textwidth]{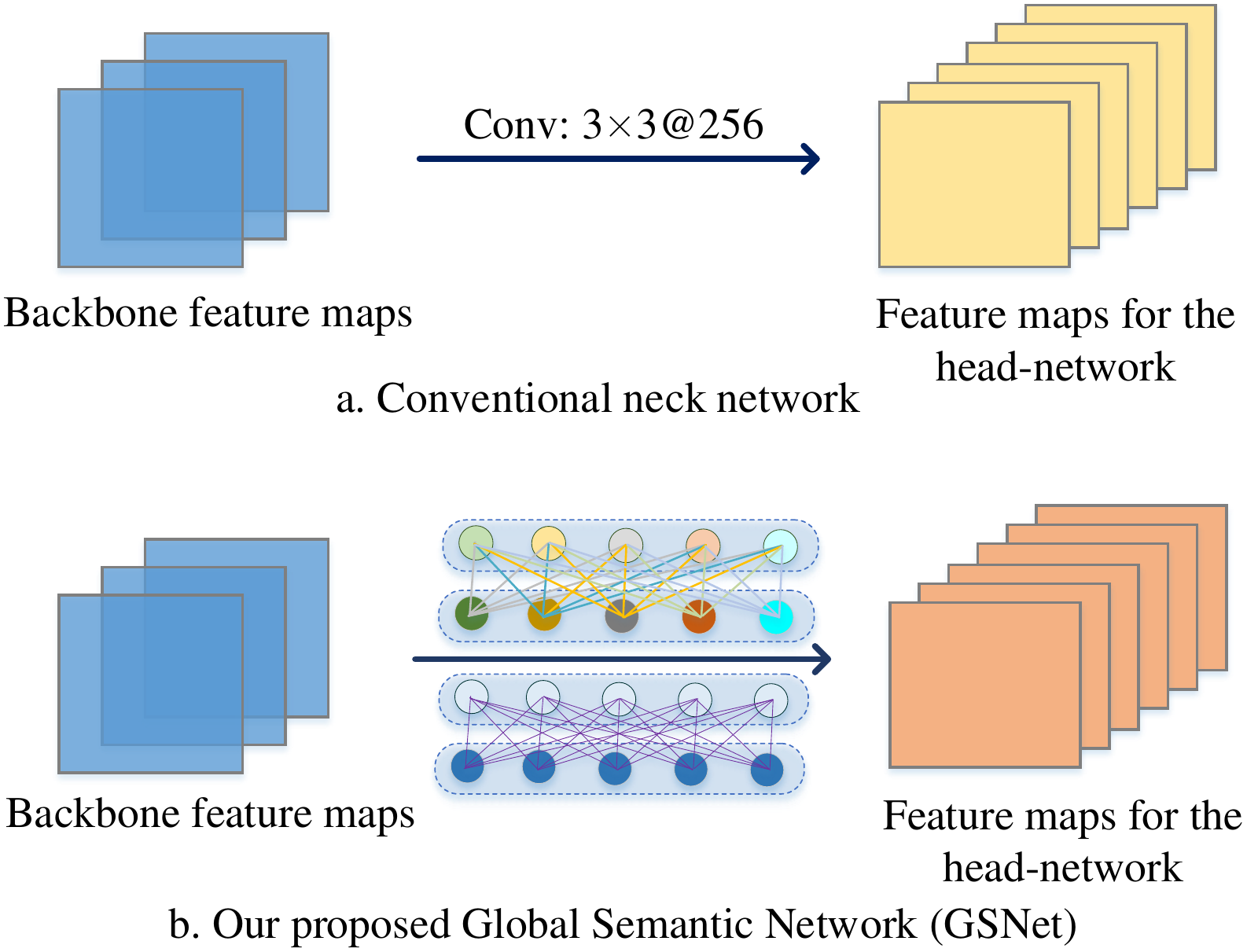}
\vspace{-3mm}
\caption{An illustration for the evolution of the neck network. The conventional neck network (a) is generally based on the coarse accumulation of multiple convolutions. Contrastively, our proposed GSNet (b) can obtain affluent backbone feature cues via a global bilateral scanning operation, thus making the model more suitable for dense prediction tasks.}
\vspace{-2mm}
\label{fig1}
\end{figure}
However, it is challenging for a standard deep CNNs model to achieve a satisfactory performance, since the ground objects in aerial images usually have the properties of tiny scale, high density, and intricate background. Especially for some overlapping and vague scenarios, the probability of an awful result is greatly increased~\cite{zhang2019positional}. In this paper, we emphasize that if the head network does not change, the reason for the unsatisfactory performance is due to the insufficient feature representation~\cite{zhang2021self,ding2019learning}. The information bottleneck mechanism~\cite{saxe2019information}, which explores the information flow between elements, can help explain this phenomenon, \ie, the imperfect neck network may cause some task-related information loss~\cite{liu2022reduce,zhao2017random}. In particular, for the non-discriminative one, the information loss problem is much more significant~\cite{zhang2018recursive,saxe2019information}.
To this end, a large number of progressive approaches are proposed to alleviate this problem which mainly start from improving the model information content~\cite{zhao2017random,chen2018encoder,guo2020augfpn,liu2018path}. For the first category of methods, \eg, random shifting~\cite{zhao2017random}, and dilated convolution~\cite{chen2018encoder}, receptive fields are expanded by adjusting the down-sampling to acquire global contexts. Methods in the second category, \eg, AugFPN~\cite{guo2020augfpn}, and PANet~\cite{liu2018path}, use multiple convolutional layers to fuse multi-scale features, such that the task-related information can be aggregated. 
 
Although some advanced head networks have also been proposed, the existing methods ignore the fact that the neck network potentially plays a pivotal role~\cite{yang2019r3det,han2021align}. As illustrated in Figure~\ref{fig1} (a), the existing neck networks generally based on the coarse accumulation of multiple convolutions, which have a limited feature aggregation ability. 
In this letter, we present a Global Semantic Network (GSNet) and a Fusion Refinement Module (FRM), which are based on the feature pyramid network~\cite{lin2017feature}. As shown in Figure~\ref{fig1} (b), GSNet can obtain the rich backbone features via a global bilateral scanning operation, thus making the model more suitable for dense prediction tasks. Besides, FRM is an active module that boosts the model representation by propagating semantic features, such that the problem of semantic gap between features at different scales can be alleviated. To demonstrate the superiority of our model, experiments are implemented on Fast R-CNN~\cite{ren2015faster} and RetinaNet~\cite{lin2017focal} on two commonly used datasets (\ie, DOTA~\cite{xia2018dota} and HRSC2016~\cite{liu2017high}) for oriented object detection. Results validate that our GSNet with FRM can achieve the top performance by $79.37\%$ and $74.49\%$ mAP for DOTA, $90.50\%$ and $90.47\%$ for HRSC2016.

Our contributions are summed up as 1) we emphasize that the information bottleneck causes the object detection network to lose information, which has a great performance damage. 2) we propose GSNet and FRM to reduce the information bottleneck by reconstructing the neck network in a bidirectional pattern. 3) our model achieves the competitive $79.37\%$ and $90.50\%$ mAP on two challenging aerial image datasets.
\begin{figure*}[t]
\centering
\includegraphics[width=.99\textwidth]{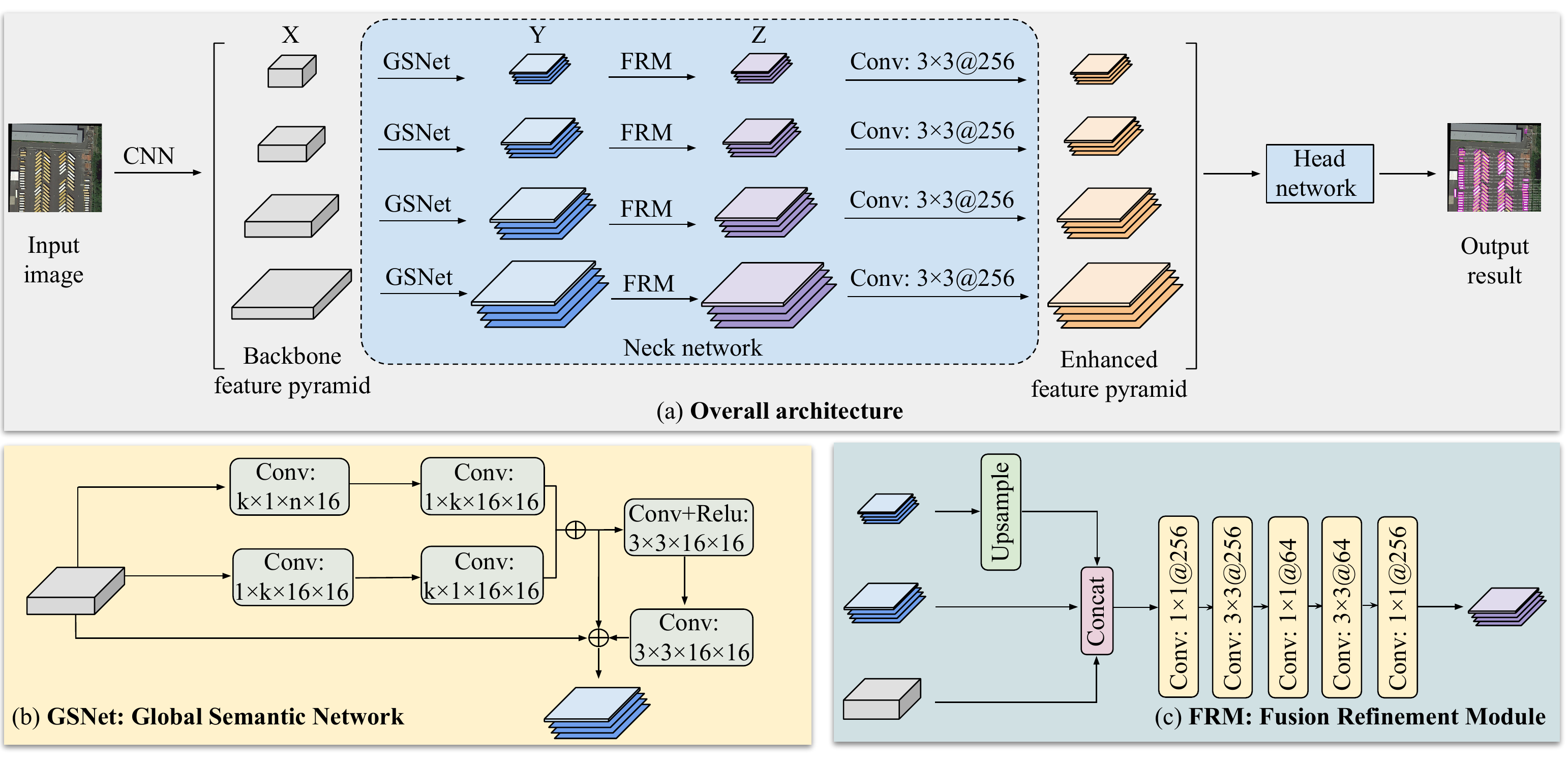}
\vspace{-2mm}
\caption{Our proposed overall architecture, where Global Semantic Network (GSNet) and Fusion Refinement Module (FRM) are implemented on each layer of the backbone feature pyramid. The detailed architectures of GSNet and FRM are shown in (b) and (c), respectively.}
\vspace{-3mm}
\label{fig2}
\end{figure*}

\section{Methodology}
\label{Ⅱ}
\subsection{Preliminaries}
Considering that the information bottleneck will cause the loss of the effective input information, while the supervised recognition model usually expects to retain the features of the input image as much as possible~\cite{zhang2018recursive,zhao2017random,guo2020augfpn,saxe2019information}. To alleviate the conflict the between information bottleneck and supervised learning models, we seek to reduce the information bottleneck to minimize feature loss and enhance the network feature representation ability.
The overall architecture is shown in Figure~\ref{fig2}. Concretely, the ImageNet~\cite{russakovsky2015imagenet} pre-trained ResNet~\cite{he2016deep} is adopted as the backbone following~\cite{ding2019learning}. Based on which, we construct the enhanced feature pyramid via the proposed  GSNet and FRM. The prediction results are finally obtained based on the enhanced feature representations with the head network, \eg, Faster R-CNN~\cite{ren2015faster} and RetinaNet~\cite{lin2017focal}.
\subsection{Global Semantic Network (GSNet)}

Compared to small convolutional layers, a large convolution kernel convolution can bring large receptive fields~\cite{peng2017large,ding2022scaling} with less computational costs, which is empirically beneficial to dense prediction tasks.
The trained classical CNNs merely identify small discriminative parts with high response, while the large effective receptive fields help recognize non-discriminative regions by sensing the high-response environment around them. Besides, large kernel enables the detection model to have a tightly connected structure that copes with different transformations.In other words, features generated by convolutions with the same kernel have a stronger spatial correlation and the fully connected layer is not suitable for localization because of its spatially sensitive nature~\cite{wu2020rethinking}. 

Motivated by the above observations, we present a GSNet in Figure~\ref{fig2} (b) that explicitly reduces feature loss and improves the model's positioning ability~\cite{peng2017large}.
First, GSNet uses as large convolution kernels as possible or even global convolutions to significantly expand the effective receptive fields.
But unlike many classification networks, GSNet does not have the large kernel convolutions of k × k directly, which would significantly increase the number of parameters. Instead, 
our GSNet employs the combined convolutions of 1 × k + k × 1 and k × 1 + 1 × k. These symmetric and depth-separable combined convolutions~\cite{peng2017large} incorporate detailed contextual information while decreasing the number of parameters and computational costs, which make it more practical.
Besides, GSNet is a fully convolutional network~\cite{long2015fully} with only linear operations applied in combined convolutions. The global bilateral scanning operation can be formulated as:
\begin{equation}
\text {M}=\operatorname{Conv}\left(\operatorname{Conv} 1 \mathrm{D}(X)^{T}\right)+\operatorname{Conv}(\operatorname{Conv} 1 \mathrm{D}(X))^{T},
\end{equation}
where X as the input are the feature maps extracted from the backbone feature pyramid.
Since the localization maps obtained by the recognition network cannot precisely represent the boundary of the target object, we refine the bounding box by modeling the boundary alignment as a residual structure to boost the accuracy. GSNet is introduced into the feature pyramid structure, which is closely linked to the feature maps and trained in an end-to-end manner, making the model more suitable for dense prediction tasks.
Formally,
\begin{equation}
\text {Y}=M+R(M)+X,
\end{equation}
\begin{equation}
R(M)=\operatorname{Conv} 2 D(\sigma(\operatorname{Conv} 2 D(M))),
\end{equation}
where R(·) is the residual branch, and $\sigma$ is the ReLU~\cite{glorot2011deep} activation function.
\subsection{Fusion Refinement Module (FRM)}
We proposed a novel FRM in Figure~\ref{fig2} (c).
Direct addition is not a reasonable approach for cross-scale fusion since feature maps from the different scales have semantic information gaps.
Compared to addition, channel-wise concatenation preserves more feature information, but it also increases the number of model parameters and computation. 
To this end, $1 \times 1$ convolutions are adopted at intervals to reduce dimension, alleviating convolution bottlenecks. 
Besides, a residual branch from the backbone is introduced to inject various spatial context information.
The residual structure superimposes depth features on the basis of the original features, realizing the fusion of global and local information.
After that, we implement stacked convolutional layers to remove the aliasing effect caused by the
interpolation, reducing information loss in the channel and enhancing the feature representational ability. The enhanced feature pyramid contains more higher-level and  semantic information. Formally, it can be expressed as
\begin{equation}\label{eq4}
\mathrm{Z}=f^{1 \times 1}\left(f^{3 \times 3}\left(f^{1 \times 1}\left(f^{3 \times 3}\left(f^{1 \times 1}\left(\left[\mathrm{X}_{i}, \mathrm{Y}_{i}, \mathrm{Y}_{i+1}\right]\right)\right)\right)\right)\right),
\end{equation}
where [·] is channel-wise concatenation, and $\mathrm{X}$, $\mathrm{Y}$ are feature maps from the backbone feature pyramid and feature maps processed by GSNet, respectively. $ i $ represents the level of the feature pyramid, which equals 1, 2, 3. $f^{1 \times 1}$ and $f^{3 \times 3}$ denote the standard $1 \times 1$ and $3 \times 3$ convolutions.

\section{Experiments}
\begin{figure}[htp]
\centering
\includegraphics[width=0.48\textwidth]{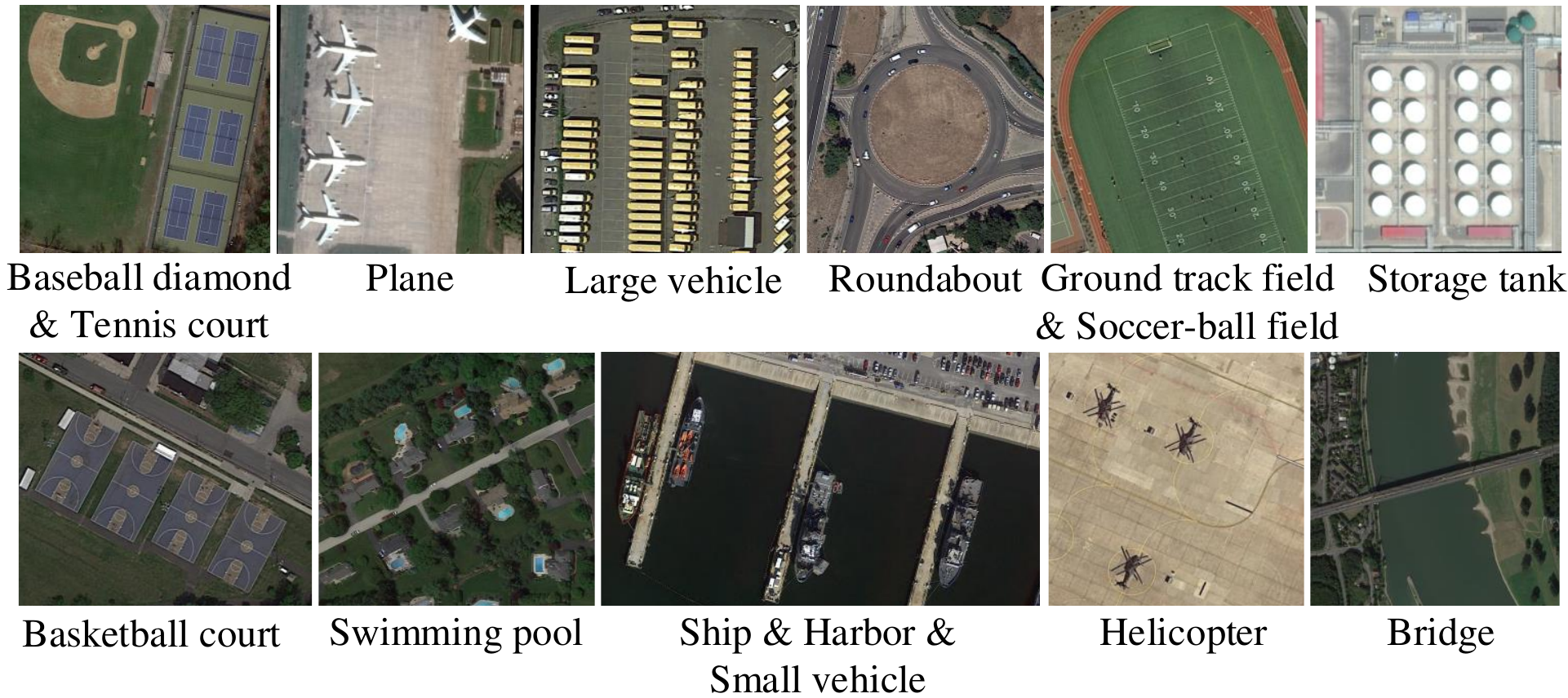}
\caption{Some demo training samples of the DOTA dataset~\cite{xia2018dota}.}
\label{fig4}
\end{figure}
\subsection{Datasets and Evaluation Metric}
{DOTA}~\cite{xia2018dota} contains 2,806 aerial images with 15 classes, whose size varies from 800 × 800 to 4000 × 4000. Figure~\ref{fig4} shows some training samples in this data set.
{HRSC2016}~\cite{liu2017high} contains 1061 images with high resolution, whose size ranges from 300 × 300 to 1500 × 900. 
For both data sets, we randomly take 3/6 for training, 1/6 and 2/6 for validation and testing, respectively. All images are cropped into $1024 \times 1024$ patches with a stride of 824.
We use mean average precision (mAP) as the primary metric. In addition, three commonly used metrics are taken into consideration to verify the model efficiency, which are the GFLOPs, the model Parameters (Params), and the Frames Per Second (FPS). 

\subsection{Experimental Setup}
\myparagraph{Baselines.} 
We deploy two-stage Faster R-CNN~\cite{ren2015faster} and one-stage RetinaNet~\cite{lin2017focal} as baseline models. ResNet101 and ResNet50 (both are pre-trained on the ImageNet~\cite{russakovsky2015imagenet}) are adopted as backbone networks. FPN is utilized to produce an enhanced neck network. 
As in~\cite{ren2015faster,lin2017focal}, the rotated head is developed in RoI-Transformer~\cite{ding2019learning} and RotatedRetinaNet~\cite{lin2017focal} individually. All experimental settings strictly follow as reported in official codes for a fair comparison. 

\myparagraph{Training Details.}
We use the standard SGD~\cite{krizhevsky2012imagenet} as the optimizer, where the learning rate is initialized to 0.005 and 0.0025 for two baselines. The weight decay and momentum are set to 0.0001 and 0.9, respectively. Models are trained for DOTA and HRSC2016 in by epochs on RTX 3060 with the batch size of 2.

\begin{table}[t]
\begin{center}
\renewcommand\arraystretch{1.3}
\setlength{\tabcolsep}{.7pt}{
\caption{Ablation study.}
\begin{tabular}{ r | c | c  c | c  c  c } 
Methods & Backbone & GSNet & FRM & mAP(\%) & Params(M) & GFLOPs \\ 
\hline \hline
Faster R-CNN~\cite{ren2015faster} & Res-101 & \xmark  & \xmark & 73.09 & 74.12 & 289.19 \\  
\cdashline{1-7}[0.8pt/2pt]
Faster R-CNN~\cite{ren2015faster} & Res-101 & \cmark  & \xmark & 
76.61$_{\color{red}{\uparrow3.52}}$ & 74.61 & 293.98\\ 
Faster R-CNN~\cite{ren2015faster} & Res-101 & \xmark  & \cmark & 78.98$_{\color{red}{\uparrow5.89}}$ & 76.45 & 325.66\\ 
Faster R-CNN~\cite{ren2015faster} & Res-101 & \cmark  & \cmark & \textbf{79.37}$_{\color{red}{\uparrow6.28}}$ & 77.91 & 338.14\\ 
\hline
RetinaNet~\cite{lin2017focal} & Res-50 & \xmark  & \xmark & 68.79 & 36.42 & 215.92 \\ 
\cdashline{1-7}[0.8pt/2pt]
RetinaNet~\cite{lin2017focal} & Res-50 & \cmark  & \xmark & 70.78$_{\color{red}{\uparrow1.99}}$ & 37.77 & 221.54 \\ 
RetinaNet~\cite{lin2017focal} & Res-50 & \xmark  & \cmark & 71.10$_{\color{red}{\uparrow2.31}}$ & 38.34 & 226.03 \\ 
RetinaNet~\cite{lin2017focal} & Res-50 & \cmark  & \cmark & \textbf{71.61}$_{\color{red}{\uparrow2.82}}$ & 39.69 & 231.66 \\ 
\end{tabular}
\label{tab1}}
\end{center}
\end{table}
\subsection{Ablation Study}
Our ablation studies aim to validate the effectiveness and efficiency of the proposed modules on different baselines and datasets. For this purpose, we conduct a series of experiments and show some visual comparisons.

\myparagraph{Effectiveness of the proposed modules.} Table~\ref{tab1} shows result comparisons for effectiveness of the proposed modules. Specifically, we take Faster R-CNN based on ResNet101 as a baseline. It is observed that GSNet and FRM improve bounding box mAP by $3.52\%$ and $5.89\%$, respectively. Combining GSNet and FRM, our method achieves $79.37\%$mAP, which is $6.28\%$ higher than the baseline by a large margin. For model efficiency, we can observe that when GSNet is implemented on the baseline, there are only 0.49M Params and 4.79 GFLOPs. It shows that the combined convolution in GSNet could effectively control model parameters and computational cost.


\myparagraph{Effectiveness on different baselines.}
Table~\ref{tab1} shows the results of our modules deployed to two baselines on DOTA~\cite{xia2018dota}. For RetinaNet, comparing row 9 to row 6, we observe that the proposed modules bring remarkable performance enhancements(\ie, $2.82\%$mAP). It is because our GSNet and FRM encourage each layer to preserve more features to reduce information bottleneck. This phenomenon is consistent across the HRSC2016 dataset~\cite{liu2017high}. 
As we mentioned under Eq.~\ref{eq4}, the main reason for computational overheads is the introduction of additional convolutional layers in constructing FRM. 

\begin{figure}[t]
\centering
\includegraphics[width=0.48\textwidth]{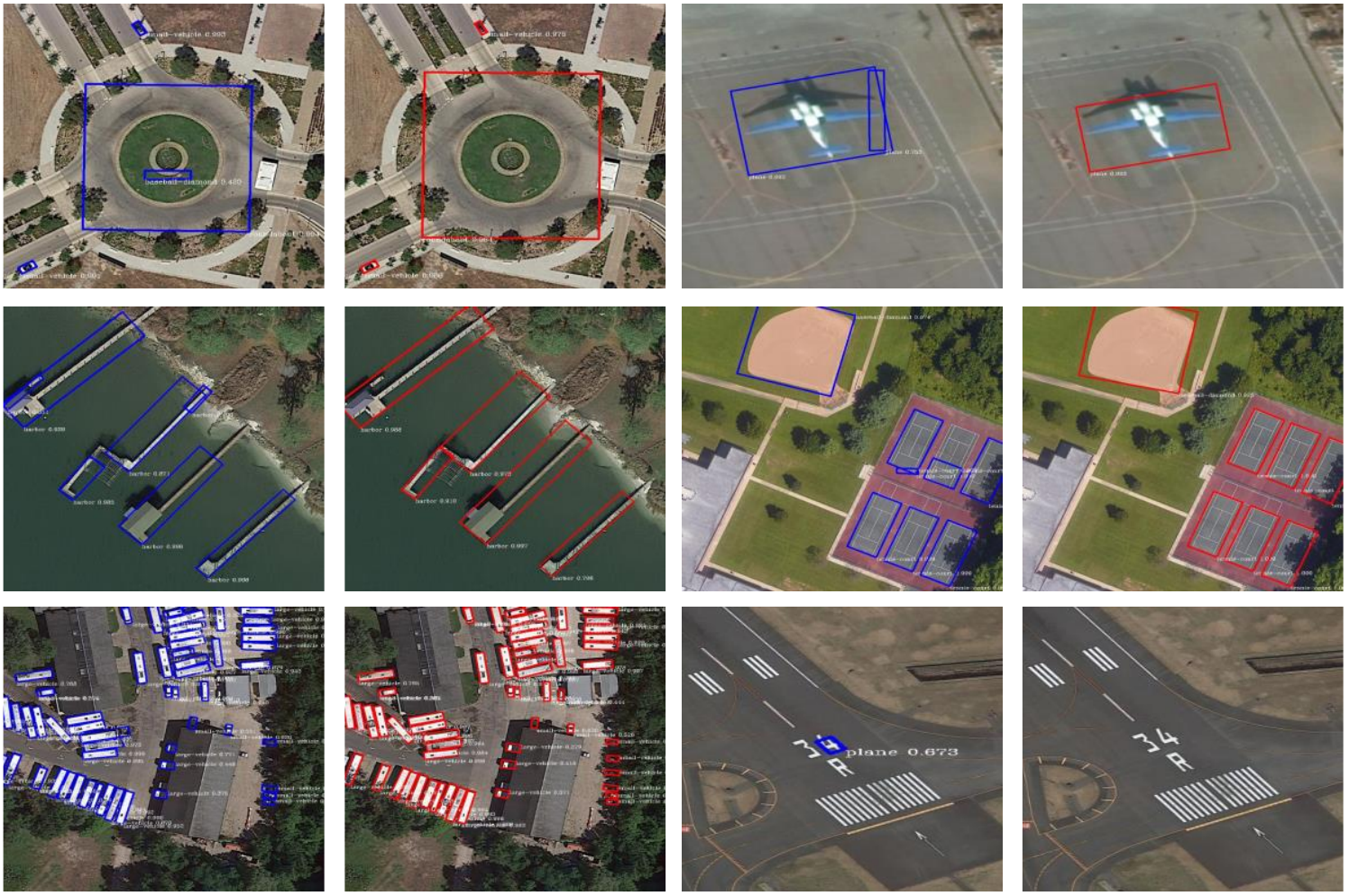}
\caption{Visualization of oriented detection results of baseline (blue boxes) and Faster R-CNN + Ours (red boxes) on DOTA dataset~\cite{xia2018dota}.}
\label{fig3}
\end{figure}
\begin{table*}[t]
\begin{center}
\renewcommand\arraystretch{1.3}
\setlength{\tabcolsep}{2.5pt}{
\caption{Result comparisons with state-of-the-art methods on DOTA dataset~\cite{xia2018dota}.}
\begin{tabular}{ r | c  c  c  c  c  c  c  c  c  c  c  c  c  c  c  c | c }
Methods & mAP(\%) & PL & BD & BR & GTF & SV & LV & SH & TC & BC & ST & SBF & RA & HA & SP & HC & FPS \\ 
\hline \hline
\emph{two-stage:} & & & & & & & & & & & & & & & & & \   \\
Gliding Vertex~\cite{xu2020gliding} & 75.02 & 89.64 & 85.00 & 52.26 & 77.34 & 73.01 & 73.14 & 86.82 & 90.74 & 79.02 & 86.81 & 59.55 & \textbf{70.91} & 72.94 & 70.86 & 57.32 & 10.0  \\
Oriented R-CNN~\cite{xie2021oriented}& 76.28 & 88.86 & 83.48 & 55.27 & 76.92 & 74.27 & 82.10 & 87.52 & 90.90 & 85.56 & 85.33 & 65.51 & 66.82 & 74.36 & 70.15 & 57.28 & \textbf{15.1} \\
CenterMap OBB~\cite{wang2020learning} & 76.03 & 89.83 & 84.41 & 54.60 & 70.25 & 77.66 & 78.32 & 87.19 & 90.66 & 84.89 & 85.27 & 56.46 & 69.23 & 74.13 & 71.56 & 66.06 & 6.3 \\
RSDet-II~\cite{qian2021learning} & 76.34 & 89.93 & 84.45 & 53.77 & 74.35 & 71.52 & 78.31 & 78.12 & \textbf{91.14} & 87.35 & 86.93 & 65.64 & 65.17 & 75.35 & 79.74 & 63.31 & -- \\
SCRDet++~\cite{yang2020scrdet++} & 76.81 & \textbf{90.05} & 84.39 & 55.44 & 73.99 & 77.54 & 71.11 & 86.05 & 90.67 & 87.32 & \textbf{87.08} & \textbf{69.62} & 68.90 & 73.74 & 71.29 & 65.08 & 13.0 \\

\cdashline{1-18}[0.8pt/2pt]
Faster R-CNN+Ours & \textbf{79.37}$_{\color{red}{\uparrow2.56}}$ & 89.66 & \textbf{86.04} & \textbf{56.25} & \textbf{79.45} & \textbf{79.07} & \textbf{84.29} & \textbf{88.40} & 90.86 & \textbf{88.10} & 85.51 & 65.56 & 66.01 & \textbf{78.70} & \textbf{79.57} & \textbf{73.02} 
& 14.0\\ 
\hline \hline
\emph{one-stage:} & & & & & & & & & & & & & & & & & \   \\
O$^2$-Det~\cite{wei2020oriented} & 71.04 & 89.31 & 82.14 & 47.33 & 61.21 & 71.32 & 74.03 & 78.62 & 90.76 & 82.23 & 81.36 & 60.93 & 60.17 & 58.21 & 66.98 & \textbf{61.03} & --\\
R$^3$Det~\cite{yang2019r3det} & 71.69 & 89.54 & 81.99 & 48.46 & 62.52 & 70.48 & 74.29 & 77.54 & 90.80 & 81.39 & 83.54 & \textbf{61.97} & 59.82 & 65.44 & 67.46 & 60.05 & 14.0\\
BBAVectors~\cite{yi2021oriented} & 72.32 & 88.35 & 79.96 & 50.69 & 62.18 & 78.43 & 78.98 & \textbf{87.94} & 90.85 & 83.58 & 84.35 & 54.13 & 60.24 & 65.22 & 64.28 & 55.70 & --\\
DRN~\cite{pan2020dynamic} & 73.23 & \textbf{89.71} & 82.34 & 47.22 & 64.10 & 76.22 & 74.43 & 85.84 & 90.57 & \textbf{86.18} & 84.89 & 57.65 & 61.93 & \textbf{69.30} & 69.63 & 58.48 & 9.8\\
CFC-Net~\cite{ming2021cfc} & 73.50 & 89.08 & 80.41 & \textbf{52.41} & \textbf{70.02} & 76.28 & 78.11 & 87.21 & \textbf{90.89} & 84.47 & \textbf{85.64} & 60.51 & 61.52 & 67.82 & 68.02 & 50.09 & -- \\
\cdashline{1-18}[0.8pt/2pt]
RetinaNet+Ours & \textbf{74.49}$_{\color{red}{\uparrow0.99}}$ & 88.92 & \textbf{82.79} & 51.93 & 69.53 & \textbf{79.13} & \textbf{79.16} & 87.26 & 90.85 & 82.19 & 85.12 & 55.34 & \textbf{66.73} & 71.28 & \textbf{70.46} & 56.64 & \textbf{20.0}
\end{tabular}
\label{tab3}}
\vspace{-6mm}
\end{center}
\end{table*}

\begin{table}[t]
\begin{center}
\renewcommand\arraystretch{1.3}
\setlength{\tabcolsep}{6 pt}{
\caption{The quantitative result comparisons with state-of-the-art methods on the test set of HRSC2016~\cite{liu2017high}.}
\begin{tabular}{ r | c | c | c }
Methods & Publication & mAP (\%) & FPS\\ 
\hline \hline
RoI Trans~\cite{ding2019learning} & CVPR 2019 & 86.20 & 6.0 \\ 
Gliding Vertex~\cite{xu2020gliding} & TPAMI 2020 & 88.20 & -- \\
R$^3$Det~\cite{yang2019r3det} &  AAAI 2021 & 89.26 & 12.0 \\  
S$^2$A-Net~\cite{han2021align} & TGRS 2021 & 90.17 & --  \\ 
\cdashline{1-4}[0.8pt/2pt]
Faster R-CNN + Ours & -- & \textbf{90.50}$_{\color{red}{\uparrow0.33}}$ & 14.0\\ 
RetinaNet + Ours & -- & \textbf{90.47}$_{\color{red}{\uparrow0.30}}$ & \textbf{25.1}
\end{tabular}
\label{tab4}}
\vspace{-6mm}
\end{center}
\end{table}
\myparagraph{Visualizations.}
Figure~\ref{fig3} shows some visual comparisons of DOTA~\cite{xia2018dota} between the baseline (blue boxes) and Faster R-CNN + Ours (red boxes). Faster R-CNN + Ours refers to our proposed network based on ResNet-101. We can intuitively observe that our proposed methods have noticeable accuracy improvement in location and boundary, \eg, the roundabout, the harbor, and the plane. From the last two rows, we observe that it also enhances the recall of some small objects.

\subsection{Comparisons with State-of-the-arts}
In this section, we make result comparisons with the state-of-the-art methods on both DOTA~\cite{xia2018dota} and HRSC2016~\cite{liu2017high}.

\myparagraph{Results on DOTA~\cite{xia2018dota}.}
As shown in Table~\ref{tab3}, compared to the previous best result of $76.81\%$ by SCRDet++~\cite{yang2020scrdet++} (\ie, the two-stage model) and $73.50\%$ by CFC-Net~\cite{ming2021cfc}(\ie, the one-stage model), our GSNet + FRM model ranks the first and improves $2.56\%$ and $0.99\%$ mAP, respectively. Concretely, some hard categories (\eg, the ship, the large vehicle, the harbor, and the helicopte) have notable mAP improvements. These results indicate that our model enhances the feature presentation capabilities.
With input image size 
of $1024 \times 1024$, our model achieves 14.0 and 20.0 FPS on two RTX 3060 GPUs, respectively. This observation can validate the efficiency of our proposed method.

\myparagraph{Results on HRSC2016~\cite{liu2017high}.}
The quantitative result comparisons on the test set of HRSC2016 are given in Table~\ref{tab3}. We can observe that our results are markedly prevail, reaching the top performance (\ie, $90.50\%$ mAP in 14.0 FPS and $90.47\%$ mAP in 25.1 FPS) among all the state-of-the-art methods, which surpass the previous best model by $0.33\%$ and $0.30\%$ mAP with comparable inference speed, individually.
\section{Conclusions}
In this letter, we first analyze the existing problems in object detection from the perspective of information bottleneck. Then, we propose a simple GSNet and FRM for enhancing feature representations for the neck network in object detection of aerial images. Extensive experiments on two challenging datasets confirm the superiority of our GSNet + FRM model.
The main limitation is that although our network greatly improves the detection rate of small targets, it still fails to detect tiny ones that are difficult to be distinguished by naked eyes.
In the future, we will consider adjusting the network structure to overcome this issue and applying our proposed methods to more computer vision tasks, \eg, semantic segmentation, video object detection.

\bibliographystyle{IEEEtran}
\bibliography{IEEE_ref}
\end{document}